\documentclass[11pt]{article}

\usepackage[final]{acl}

\usepackage{times}
\usepackage{latexsym}
\usepackage{enumitem}

\usepackage[T1]{fontenc}

\usepackage[utf8]{inputenc}
\usepackage{placeins} 

\usepackage{microtype}

\usepackage{inconsolata}

\usepackage{graphicx}

\usepackage{xcolor}

\usepackage[ruled,vlined]{algorithm2e}
\usepackage{tabularx}
\usepackage{tcolorbox}
\tcbuselibrary{breakable}

\usepackage{minted}

\title{VIGIL: Towards Edge-Extended Agentic AI for Enterprise IT Support}

\author{
\textbf{Sarthak Ahuja\textsuperscript{*} \quad
Neda Kordjazi\textsuperscript{*} \quad
Evren Yortucboylu\textsuperscript{*} \quad
Vishaal Kapoor \quad
Mariam Dundua} \\
\textbf{Yiming Li \quad
Derek Ho \quad
Vaibhavi Padala \quad
Jennifer Whitted \quad
Rebecca Steinert} \\
Amazon Engine, AI Center of Excellence \\
\texttt{\{sarahuja,nedakord,yortuc,vishaalk,madundua\}@amazon.com} \\
\texttt{\{yimingll,derekjho,vapadala,jenwhitt,rsteinrt\}@amazon.com} \\
\textsuperscript{*}Equal contribution
}

\begin{document}
\maketitle

\begin{abstract}
Enterprise IT support is constrained by heterogeneous devices, evolving
policies, and long-tail failure modes that are difficult to resolve
centrally. We present VIGIL, an edge-extended agentic AI system that
deploys desktop-resident agents to perform situated diagnosis, retrieval
over enterprise knowledge, and policy-governed remediation directly on
user devices with explicit consent and end-to-end observability. In a
10-week pilot of VIGIL's operational loop on 100 resource-constrained endpoints, VIGIL reduces
interaction rounds by 39\%, achieves at least 4$\times$ faster
diagnosis, and supports self-service resolution in 82\% of matched
cases. Users report excellent usability, high trust, and low cognitive
workload across four validated instruments, with qualitative feedback
highlighting transparency as critical for trust. Notably, users rated
the system higher when no historical matches were available, suggesting
on-device diagnosis provides value independent of knowledge base
coverage. This pilot establishes safety and observability foundations for fleet-wide continuous improvement.


\end{abstract}

\section{Introduction}

Large enterprise IT ecosystems comprise vast fleets of heterogeneous devices, networks, and policies operating under partial observability \cite{li2021fog}. Failures often emerge from subtle local interactions rather than clear global faults, making diagnosis reactive and slow. Operational knowledge remains fragmented, and support workflows continue to rely on centralized human intervention and static playbooks, with AI chatbots offering limited post-incident assistance \cite{reinhard2024_genai_it_support}. 


In parallel, AI systems are evolving beyond conversational assistants toward agentic, computer-use models capable of acting within software environments through terminal interfaces and operational tools \cite{claude_code_overview,kiro_cli_intro,shen2024taskbench,zhuang2023toolqa}. Enterprise platforms are beginning to incorporate such capabilities into structured IT service workflows \cite{zhang2025agenticai_servicenow,servicenowAccenture2025whitepaper}, typically through centrally orchestrated agents within predefined automation boundaries rather than distributed, on-device execution across heterogeneous endpoints. Executing bounded actions directly on edge devices can enable lower-latency remediation, access to fine-grained local context, and improved resilience under partial connectivity, but such distributed autonomy requires mechanisms that prevent out-of-bounds behavior.




VIGIL explores this opportunity by bringing agentic and computer-use capabilities into enterprise IT through an emerging edge-extended architecture that combines on-device autonomy with cloud coordination and continuous improvement. At its core, VIGIL relies on language understanding, retrieval-augmented generation, and natural language reasoning to interpret user-reported issues, ground diagnosis in enterprise knowledge, and communicate remediation steps transparently.
VIGIL treats each endpoint as an intelligent participant running specialized agents for diagnosis, knowledge access, and remediation, while cloud services provide model inference, fleet-level observability, and structured escalation to human operators.


 Each interaction produces structured experience signals that form the basis for future learning at the context and prompt levels without requiring model retraining. This paper makes the following contributions:
\begin{itemize}
    \item A distributed architecture for enterprise IT support in
    which endpoints perform situated diagnosis and remediation under
    cloud-coordinated governance and observability
    \item An edge-device controller that integrates on-device diagnostic and remediation agents with enterprise knowledge access, observability, and structured escalation to human operators
    \item A proof-of-value pilot evaluating the edge controller component on real IT support scenarios, providing early evidence on efficiency, usability, and user trust.
\end{itemize}

\section{Related Work}

\paragraph{LLM-enhanced IT operations.}
Recent work demonstrates that large language models can enhance AIOps by structuring logs and incidents for improved detection and decision-making \cite{vitui2025empowering}, and surveys document both the promise of such techniques and the persistence of centralized, human-mediated workflows \cite{li2025aiops_survey}. Agent-oriented efforts explore partial autonomy in detection, classification, and remediation \cite{zota2025agentic}, while evaluation frameworks probe the reliability of LLM-based operational agents \cite{chen2025aiopslab}. However, these systems retain centralized architectures in which endpoints serve as passive telemetry sources rather than active participants in diagnosis and resolution.

\paragraph{Agentic and tool-augmented systems.}
The broader agentic AI landscape has advanced rapidly, with surveys characterizing the design space of autonomous agents and their orchestration patterns \cite{acharya2025agentic,sapkota2025ai}. Tool-augmented language models enable structured interaction with external systems through learned or specified tool interfaces \cite{schick2023toolformer,patil2023gorilla}, and standardized protocols such as MCP \cite{MCP} and A2A \cite{a2a2025spec} provide interoperability across agent boundaries. Retrieval-augmented generation grounds model outputs in curated knowledge \cite{gao2023rag_survey}. VIGIL builds on these capabilities but is the first to integrate them into a governed, edge-deployed IT operations system.

\paragraph{Edge and distributed AI.}
Recent proposals advocate distributing intelligence across edge devices rather than concentrating it in centralized services \cite{luo2025toward,tallam2025autonomous}. While these works establish architectural principles for multi-LLM edge systems, they remain largely theoretical and do not address the governance, observability, and safety constraints required for enterprise IT. VIGIL addresses this gap by operating under deterministic policy control.
\section{VIGIL Vision and Architecture}

\begin{figure}[t]
  \centering
  \includegraphics[width=1.0\columnwidth]{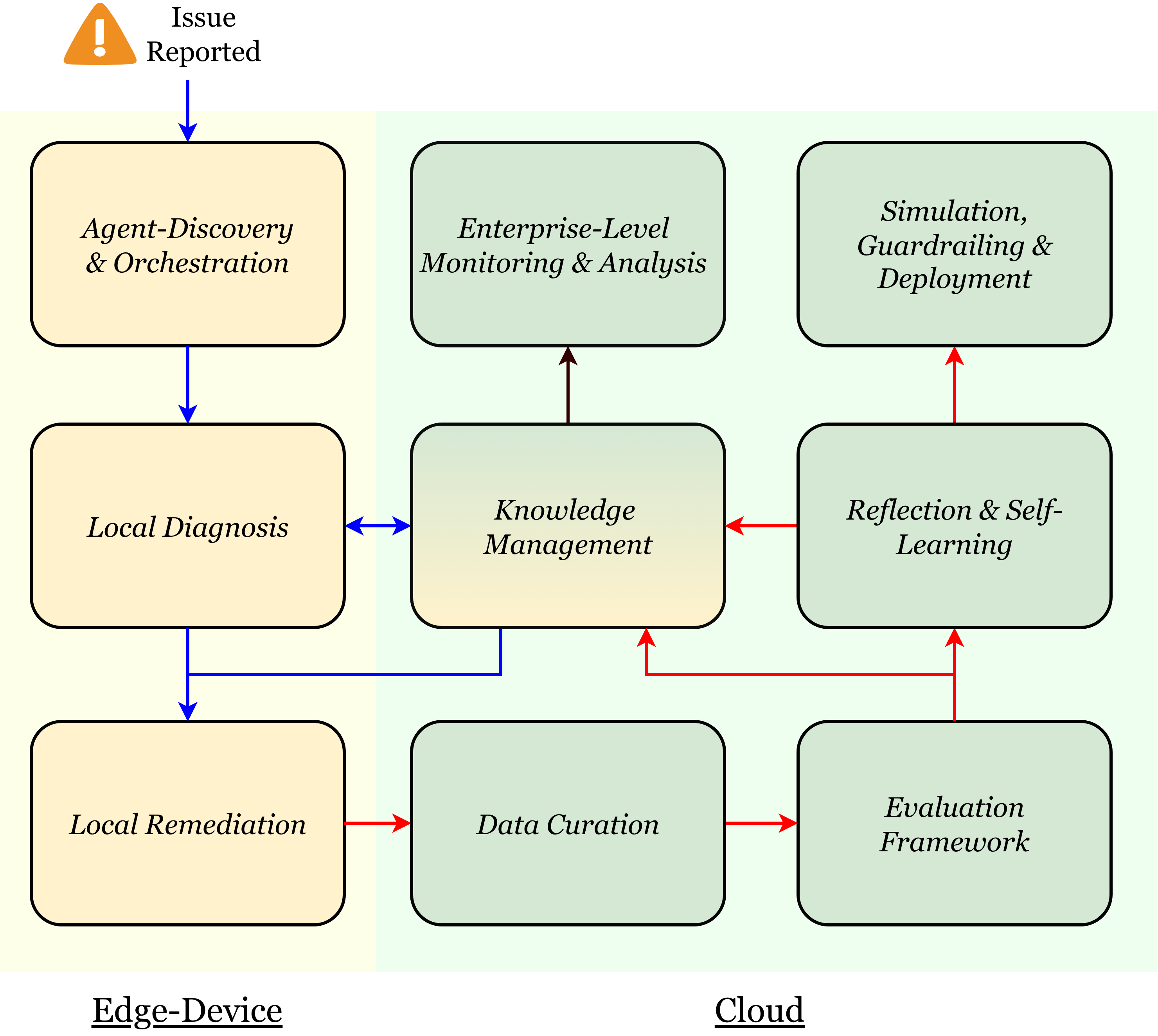}
  \caption{Operational and self-improvement loops within VIGIL. Device-level cognition, the focus of this paper, executes locally on endpoints (shown in {\color{blue}\textbf{blue}}), while cloud-level reflection, designed for future activation, refines policies, memory, and prompts over time (shown in {\color{red}\textbf{red}}). Monitoring at the fleet level operates in parallel across both loops (shown in {\color{black}\textbf{black}}).}
  \label{fig:flow}
\end{figure}

VIGIL is organized around two tightly coupled control loops that together define how intelligence, action, and adaptation are distributed across the enterprise (depicted in figure \ref{fig:flow}). 

\paragraph{Operational loop.}
The operational loop governs real-time diagnosis and remediation directly on endpoint devices, and is the primary focus of this paper. 
On each endpoint, the loop follows a structured diagnose, retrieve, and remediate workflow: when anomalies are detected through telemetry, scheduled checks, or user-reported issues, the system collects evidence, retrieves relevant enterprise knowledge, and executes bounded remediation actions. All steps are logged as structured traces for downstream analysis and monitoring. Execution is resilient to partial connectivity, allowing devices to continue troubleshooting using cached knowledge and locally enforced policies even when cloud services are unavailable.

\begin{figure*}[th]
  \centering
  \includegraphics[width=0.95\textwidth]{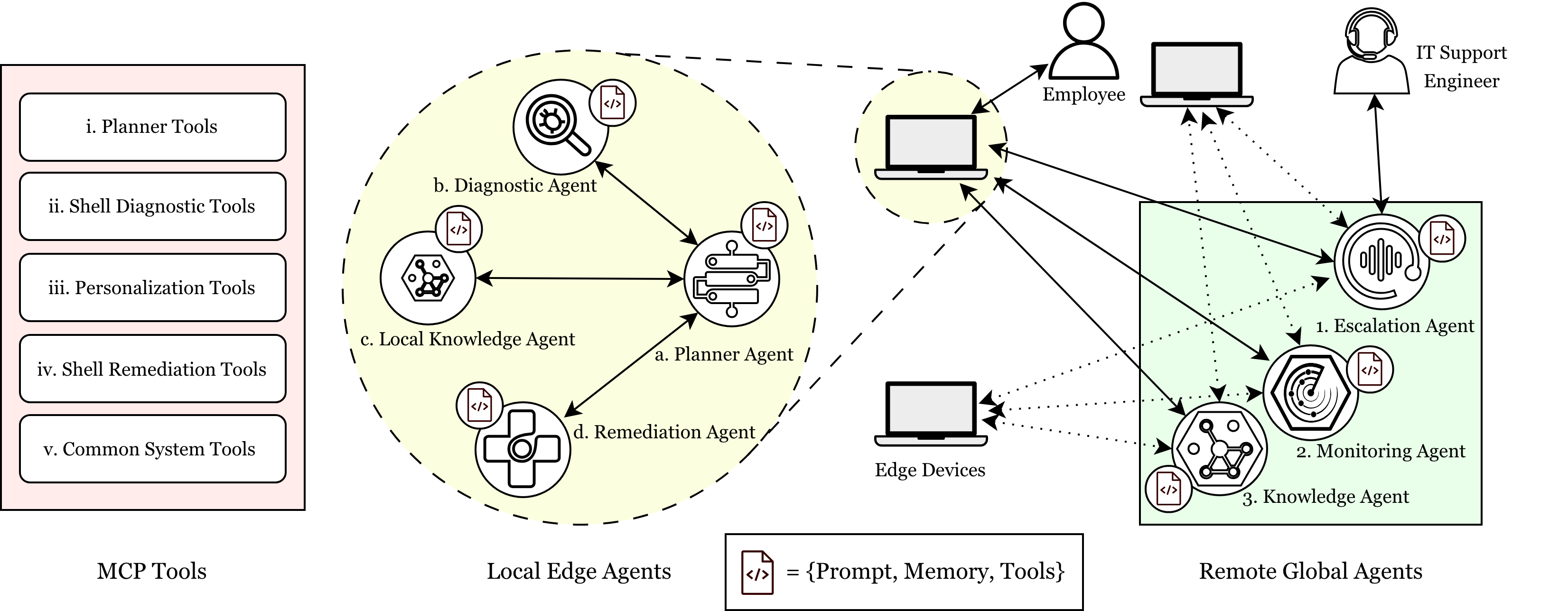}
  \caption{VIGIL’s agentic architecture integrating local edge agents and remote cloud agents. Edge agents execute diagnosis and remediation on-device, while cloud agents provide coordination, governance, and observability across the fleet.}
  \label{fig:agents}
\end{figure*}

\paragraph{Self-improvement loop.}
The self-improvement loop operates over structured experience produced by the operational loop. Rather than retraining foundation models, it refines the contextual scaffolding that shapes agent behavior, including memory \cite{ReasoningBank}, retrieval strategies \cite{gao2023rag_survey}, prompts and workflows \cite{GEPA}, and governance policies \cite{santurkar2023whose}. This loop reflects the broader VIGIL vision and is left as future work, to be activated once sufficient safety and observability guarantees are established.



\subsection{Agent Roles}

 VIGIL realizes the operational loop through a small set of coordinated agents running on each endpoint (figure \ref{fig:agents}). These agents are tightly integrated components of a single control system, sharing state and operating over constrained, auditable tool interfaces \cite{yao2022react,schick2023toolformer,patil2023gorilla,wang2023longmem}.

\paragraph{Planner Agent.}
The planner agent orchestrates the local workflow. It decomposes troubleshooting goals into diagnostic steps and remediation plans, employing structured reasoning methods such as ReAct-style deliberation and multi-step reasoning \cite{yao2023react,yao2023tree}. The planner reasons over device context, episodic history, and enterprise policies, delegating evidence collection and actuation to specialized agents while keeping decision logic explicit and governable \cite{a2a2025spec}.

\paragraph{Diagnosis Agent.}
The diagnosis agent gathers structured evidence about local system state using tools exposed through the Model Context Protocol (MCP). Tool usage is constrained by purpose, preconditions, and safety annotations, consistent with modern tool-augmented language model systems \cite{schick2023toolformer,patil2023gorilla}. Probe selection is informed by uncertainty-aware debugging and information-gain-driven exploration \cite{kirsch2022minimal,agrawal2024deepuncertainty,gupta2023llmdebugger,moshkovich2025taming}. The output is a structured diagnostic profile.

\paragraph{Knowledge Agent.}
The knowledge agent manages contextual information used throughout the loop. Locally, it maintains compact episodic memory capturing device-specific history. Remotely, it interfaces with curated enterprise knowledge and federated summaries from other devices. Together, these form a hybrid memory substrate integrating episodic, semantic, and causal structure \cite{anokhin2024arigraph,cai2025kgllm,fiorini2025episodic}.

\paragraph{Remediation Agent.}
The remediation agent executes bounded, reversible interventions such as service restarts, configuration changes, or rollbacks. Actions are validated against deterministic safety and compliance constraints, gated by user consent. Execution follows stepwise verification and self-correction paradigms drawn from recent work on automated repair and safe actuation \cite{Bouzenia2025RepairAgent,santurkar2023whose,madaan2023selfrefine,zhu2023repairagent}. All actions produce auditable traces.

While execution remains local, the VIGIL design includes cloud-based agents for fleet-level coordination, observability, and governance, enabling cross-device pattern discovery \cite{laptev2023multivariate,yu2022graphanomaly} and structured escalation to human operators \cite{miller2019explanation}.

\section{Implemented Edge System}

VIGIL’s deployed edge layer is implemented as a standalone desktop application built on top of Strands \cite{strands_docs2024} and Bedrock \cite{bedrock2024} that executes diagnosis, reasoning, and remediation directly on endpoint devices. 



Upon a user-reported issue, the system first invokes a bounded set of operating-system-specific diagnostic tools exposed through the Model Context Protocol (MCP) \cite{MCP} to collect evidence and produce a structured diagnostic profile (refer Table \ref{tab:mcp}. Based on this profile, the system performs retrieval-augmented reasoning over two enterprise-curated sources: a knowledge base of IT portal articles and a repository of previously resolved cases authored by support engineers. Retrieved artifacts are assembled into a grounded context package, from which a stepwise remediation plan is synthesized and executed locally under policy control and continuous verification.

\begin{table}[th]
\centering
\small
\begin{tabular}{p{0.30\linewidth} p{0.57\linewidth}}

\textbf{Tool} & \textbf{Purpose} \\
\hline
\\[-6pt]
\texttt{system\_uptime} & Retrieves device uptime to identify reboots\\
\texttt{security\_updates} & Detects pending enterprise security updates \\
\texttt{cpu\_process} & Retrieves top running processes by CPU usage \\
\texttt{disk\_usage} & Reports disk usage across all mount points. \\
\texttt{network\_status} & Retrieves network connectivity status \\
\end{tabular}
\caption{Illustrative diagnostic tools used in the diagnose--retrieve--remediate loop.}
\label{tab:mcp}
\end{table}

All state-changing actions are mediated locally by a deterministic policy engine based on Open Policy Agent (OPA) \cite{opa2024}. Generated commands are evaluated against declarative policies and classified into one of three tiers: allow, warn, or deny. Low-risk actions execute automatically; moderate-risk actions require explicit user consent with a natural-language explanation of purpose and impact; and high-risk actions are blocked. Remediation is executed incrementally with verification after each step; if health signals regress, the system adapts its plan or escalates, and where possible selects reversible actions to enable rollback.

\section{Evaluation}
\label{sec:eval}

We conducted a proof-of-value (PoV) pilot to evaluate VIGIL under real enterprise operating conditions. The evaluation combines LLM-based automated assessment of operational effectiveness with direct human evaluation through validated user experience instruments.

\subsection{Deployment Setting}

The pilot was conducted over a 10-week period across 100 enterprise endpoints running a single hardware and software configuration (Windows-based HP G8 devices). This device category was selected because it is among the highest drivers of IT support contacts in the organization: the devices are resource-constrained and frequently exhibit issues related to the proprietary enterprise software stack required on each machine. Participants were instructed to use VIGIL as their first point of contact for IT issues they would otherwise escalate to human support, and all interaction traces were recorded for subsequent analysis.

\subsection{Operational Trace Analysis}

To quantify VIGIL's impact relative to conventional IT support, we performed a retrospective case-matching analysis against a centralized graph repository (CGR) containing over 60{,}000 pre-resolved IT support interactions authored by human support engineers. Each VIGIL session was matched to historically similar CGR cases using a two-step process: semantic similarity via dense embeddings (threshold 0.55), followed by language-model verification of full case details (confidence $\geq$ 7/10). This yielded 826 confirmed matches spanning 60 of 153 VIGIL sessions (39\%). For matched cases, we compared interaction efficiency and response time, assessed response quality across five dimensions (issue understanding, root cause accuracy, relevance, actionability, completeness) using automated evaluation, and estimated self-service potential.


\subsection{User Experience Evaluation}

At the conclusion of the pilot, participants completed a structured questionnaire comprising four validated instruments:
\begin{itemize}
    \item \textbf{System Usability Scale (SUS)} \cite{brooke1996sus}: a 10-item Likert scale providing a composite usability score (0--100), with an established industry average of 68.
    \item \textbf{NASA Task Load Index (NASA-TLX)} \cite{hart1988nasa}: a six-dimensional workload assessment covering mental demand, physical demand, temporal demand, performance, effort, and frustration.
    \item \textbf{Trust in Automation} \cite{jian2000trust}: a scale measuring user confidence in automated system decision-making and reliability.
    \item \textbf{Technology Acceptance Model (TAM)} \cite{davis1989tam}: items assessing perceived usefulness, perceived ease of use, and behavioral intention to adopt.
\end{itemize}
All reverse-scored items were handled according to each instrument's standard scoring procedure. Of the 100 pilot participants, 23 completed the full survey. Respondents also provided optional open-ended feedback on their experience.

\section{Results and Discussion}
We report results along two complementary axes: operational effectiveness of the edge component measured through trace analysis against historical IT support records, and user experience assessed through validated survey instruments.

\subsection{Operational Effectiveness}

Table~\ref{tab:operational} summarizes key metrics from the case-matching analysis described in Section~\ref{sec:eval}. Of 153 VIGIL sessions recorded during the pilot, 60 (39\%) were matched to historically similar cases in the centralized graph repository.


\begin{table}[t]
\centering
\small
\begin{tabularx}{\columnwidth}{>{\raggedright\arraybackslash}X c c}
\textbf{Metric} & \textbf{SHIELD} & \textbf{Human Support} \\
\hline
\\[-6pt]
Interaction rounds (median) & 11 cycles & 18 turns\textsuperscript{$\ast$} \\
Diagnosis time (median) & 36.5\,s & $\geq$2.7\,min\textsuperscript{$\dagger$} \\
Response quality (median) & 8.0\,/\,10 & --- \\
\hline
\\[-6pt]
Self-service potential & \multicolumn{2}{c}{82\% (high + medium confidence)} \\
Tool success rate & \multicolumn{2}{c}{95.3\% across 1{,}586 calls} \\
\end{tabularx}
\caption{Operational metrics on matched cases (n=60). 
\textsuperscript{$\ast$}Human support rounds reflect full conversational turns between user and support engineer. Interaction rounds were reduced by 39\% (83\% of cases).
\textsuperscript{$\dagger$}Conservative lower bound assuming 10\% of median contact duration (26.5\,min) reflects active work; realistic estimates yield 4--17$\times$ speedup.}
\label{tab:operational}
\end{table}

\paragraph{Efficiency.} VIGIL required a median of 11 autonomous reasoning cycles to produce a complete diagnosis, compared to 18 back-and-forth exchanges between users and human agents for matched cases, a 39\% reduction observed in 50 of 60 matched pairs. VIGIL's cycles are internal (tool invocation, reasoning, verification) and require no user interaction beyond the initial problem statement. Median diagnosis time was 36.5\,seconds; for example, when a user reported persistent application crashes, the diagnosis agent identified memory pressure from a background process, retrieved a matching resolution, and executed a targeted service restart within this window. Even conservatively assuming only 10\% of recorded human contact time (median 26.5\,minutes) is active diagnostic work, VIGIL is at least 4$\times$ faster.


\paragraph{Response quality.} Automated evaluation across five dimensions- issue understanding (8.5), response relevance (8.1), actionability (7.8), completeness (7.4), and root cause accuracy (7.1), yielded a median overall score of 8.0\,/\,10, with 81\% of sessions rated Good or Excellent. Resolution confidence analysis estimated that 82\% of cases could be resolved through self-service (39\% high confidence, 43\% medium), suggesting that the majority of issues handled by VIGIL would not require human escalation. Root cause accuracy was the lowest-scoring dimension, indicating an area for targeted improvement through richer diagnostic tooling and knowledge coverage.

\paragraph{Operational reliability.} Across all 153 sessions, VIGIL invoked diagnostic tools in 95.4\% of cases with a 95.3\% tool execution success rate over 1{,}586 total calls, confirming that the MCP-based tool interface operates reliably under real enterprise conditions.

\subsection{User Experience}
\begin{figure}
        \centering
        \includegraphics[width=1\linewidth]{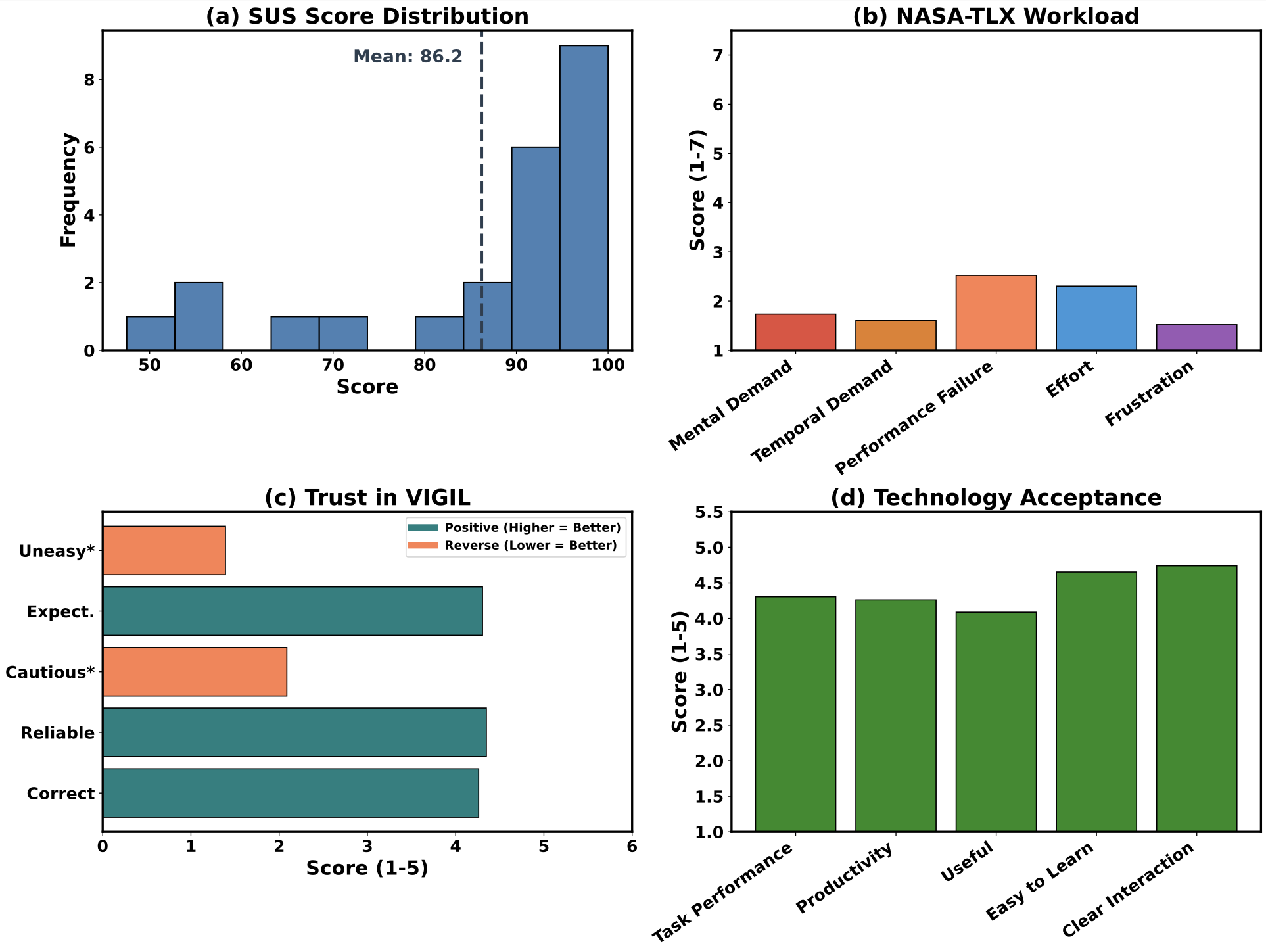}
        \caption{User experience survey: (a) SUS score distribution, (b) NASA-TLX workload, (c) user trust (bottom-left) and (d) technology acceptance}
        \label{fig:ux_detail}
    \end{figure}

\begin{figure}[th]
    \centering
    \includegraphics[width=0.8\linewidth]{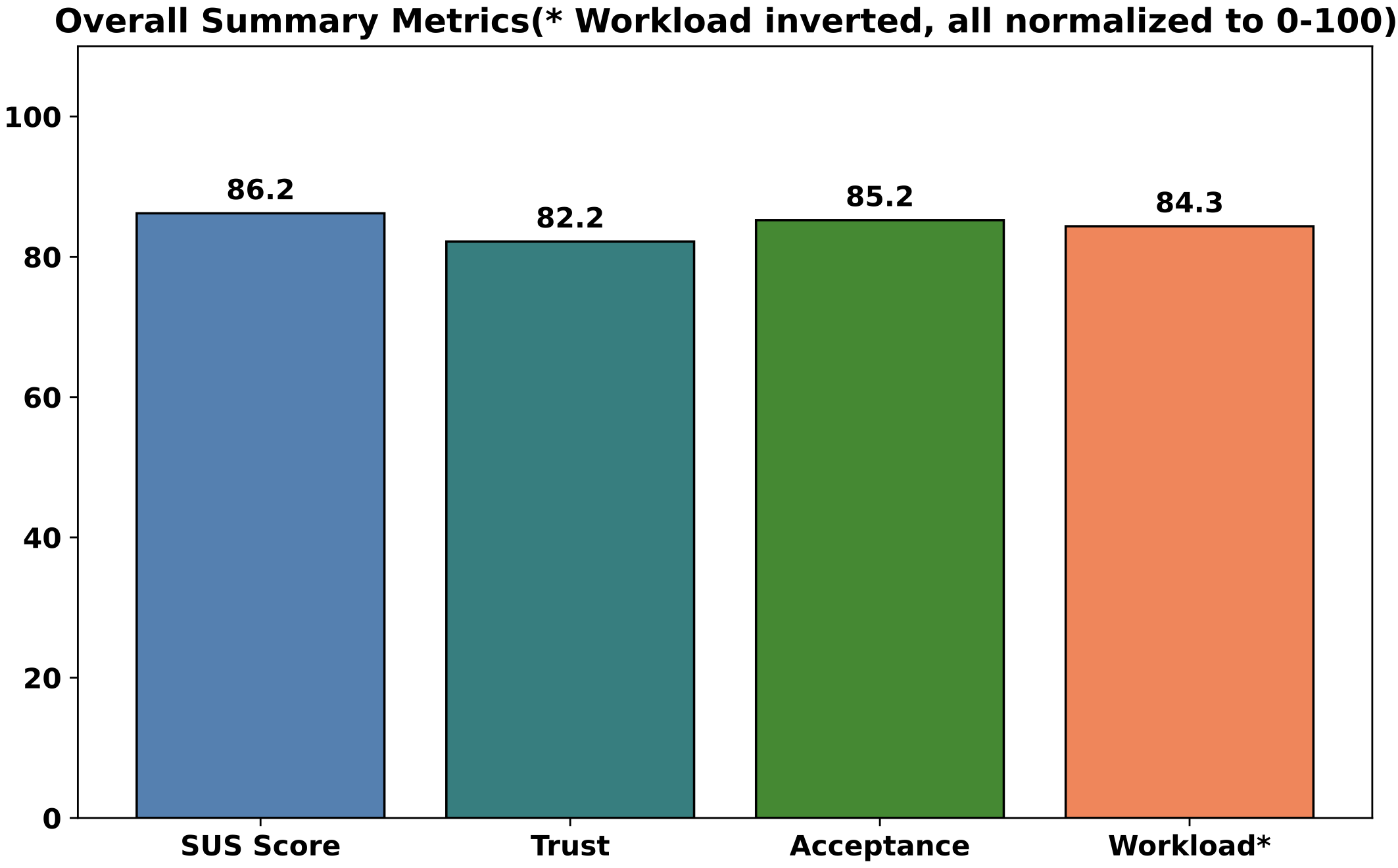}

        \caption{Normalized summary of the user experience survey across the four validated instruments (n=23)}
    \label{fig:survey_results}
\end{figure}

Table~\ref{tab:survey} and Figure~\ref{fig:survey_results} summarize responses from 23 participants who completed the post-pilot questionnaire.

\begin{table}[t]
\centering
\small
\begin{tabular}{l c}
\textbf{Instrument} & \textbf{Score} \\
\hline
\\[-6pt]
SUS & 86.2\,$\pm$\,15.7 \\
Trust in Automation & 4.29\,/\,5.0 \\
TAM & 4.41\,/\,5.0 \\
NASA-TLX & 2.53\,/\,7.0 \\
\end{tabular}
\caption{User experience scores across four validated instruments (n=23). SUS industry average is 68.0.}
\label{tab:survey}
\end{table}


As shown in Figure~\ref{fig:ux_detail}, the SUS score of 86.2 places VIGIL 18 points above the industry average of 68.0, in the Excellent tier corresponding to the 90th percentile of evaluated systems \cite{brooke1996sus}.High trust (4.29/5.0) and technology acceptance (4.41/5.0) scores indicate that users found VIGIL's recommendations reliable and believed it improved their productivity. The low NASA-TLX workload score (2.53/7.0) confirms that the system reduces rather than adds cognitive burden, a common concern with AI assistance tools.

\paragraph{Situated diagnosis exceeds retrieval alone.} A stratified analysis revealed that the 16 respondents who experienced sessions without matching historical cases rated VIGIL higher across all instruments (SUS 89.2, Trust 4.44, TAM 4.55, TLX 2.29) compared to the overall population. This suggests that VIGIL's on-device diagnostic capabilities and transparent reasoning provide value independent of knowledge base coverage, supporting the architectural decision to invest in situated diagnosis rather than relying solely on retrieval.

\paragraph{Qualitative feedback.} Open-ended responses (n=8) highlighted three themes: (1)~users valued the transparency of VIGIL's actions and reasoning as critical for trust; (2)~several requested elevated permissions for autonomous task completion, indicating demand for broader actuation scope; and (3)~multiple respondents requested continued access and wider rollout.

\subsection{Limitations}

Several factors qualify these results. The case-matching analysis covers only 39\% of VIGIL sessions; unmatched cases may differ systematically in complexity. Response quality was assessed via automated LLM-based evaluation rather than human expert judgment. The survey response rate (23 of 100 participants) introduces potential self-selection bias. Finally, the pilot was conducted on a single device category; generalization to heterogeneous fleets remains to be validated.

\section{Conclusion}

We presented VIGIL, an edge-extended agentic AI architecture for enterprise IT support, and evaluated its on-device diagnosis and remediation component through a real-world pilot. Across 100 resource-constrained endpoints over 10 weeks, the system reduced interaction rounds by 39\%, achieved at least 4$\times$ faster diagnosis, and demonstrated high response quality, user trust, and low cognitive workload. Notably, users rated the system higher when no historical matches were available, indicating that on-device reasoning adds value beyond knowledge base lookup.

By colocating reasoning and actuation with the environment where failures occur, VIGIL reduces latency, preserves fine-grained context, and enables resilient operation under partial connectivity. At the same time, distributing autonomy to heterogeneous endpoints introduces architectural challenges in governance, monitoring, and deciding when local adaptations should generalize across the fleet. Although VIGIL is designed to support improvement through context refinement and prompt optimization rather than model retraining, these mechanisms were not exercised in the current deployment. Overall, our findings suggest that distributing agentic intelligence to the edge under deterministic policy control offers a viable architectural direction for scalable, governed autonomy in enterprise IT operations.

\section*{Ethical Considerations}

All 100 pilot participants provided informed consent prior to enrollment. Interaction traces collected during the pilot were anonymized before analysis, with no personally identifiable information retained. The system's policy engine ensures that all state-changing actions on user devices are governed by deterministic safety constraints, with moderate-risk actions requiring explicit user approval and high-risk actions blocked unconditionally. No actions are executed without user awareness, and all agent reasoning and actions are logged transparently for auditability.

\bibliography{references,anthology}

@inproceedings{li2021fog,
  author    = {Liqun Li and Xu Zhang and Xin Zhao and Hongyu Zhang and Yu Kang and Pu Zhao and Bo Qiao and Shilin He and Pochian Lee and Jeffrey Sun and Feng Gao and Li Yang and Qingwei Lin and Saravanakumar Rajmohan and Zhangwei Xu and Dongmei Zhang},
  title     = {Fighting the Fog of War: Automated Incident Detection for Cloud Systems},
  booktitle = {2021 USENIX Annual Technical Conference (USENIX ATC 21)},
  year      = {2021},
  month     = jul,
  pages     = {131--146},
  publisher = {USENIX Association},
  isbn      = {978-1-939133-23-6},
  url       = {https://www.usenix.org/conference/atc21/presentation/li-liqun}
}

@inproceedings{zhuang2023toolqa,
  title     = {ToolQA: A Dataset for LLM Question Answering with External Tools},
  author    = {Zhuang, Yuchen and Yu, Yue and Wang, Kuan and Sun, Haotian and Zhang, Chao},
  booktitle = {Advances in Neural Information Processing Systems},
  year      = {2023},
  volume    = {36},
  note      = {NeurIPS 2023 Datasets and Benchmarks Track},
  url       = {https://proceedings.neurips.cc/paper_files/paper/2023/hash/9cb2a7495900f8b602cb10159246a016-Abstract-Datasets_and_Benchmarks.html},
  eprint    = {2306.13304},
  archivePrefix = {arXiv},
  doi       = {10.48550/arXiv.2306.13304}
}

@inproceedings{shen2024taskbench,
  title     = {TaskBench: Benchmarking Large Language Models for Task Automation},
  author    = {Shen, Yongliang and Song, Kaitao and Tan, Xu and Zhang, Wenqi and Ren, Kan and Yuan, Siyu and Lu, Weiming and Li, Dongsheng and Zhuang, Yueting},
  booktitle = {Advances in Neural Information Processing Systems},
  year      = {2024},
  volume    = {37},
  note      = {NeurIPS 2024 Datasets and Benchmarks Track},
  url       = {https://proceedings.neurips.cc/paper_files/paper/2024/hash/085185ea97db31ae6dcac7497616fd3e-Abstract-Datasets_and_Benchmarks_Track.html},
  eprint    = {2311.18760},
  archivePrefix = {arXiv},
  doi       = {10.48550/arXiv.2311.18760}
}

@misc{zhang2025agenticai_servicenow,
  title        = {Agentic AI (AI Agent) Development Guidelines and Use Cases (Hands-on Experience)},
  author       = {Zhang, John},
  howpublished = {ServiceNow Community (Blog)},
  year         = {2025},
  month        = mar,
  url          = {https://www.servicenow.com/community/in-other-news/agentic-ai-ai-agent-development-guidelines-and-use-cases-hands/ba-p/3206822},
  note         = {Accessed 2026-02-11}
}

@misc{servicenowAccenture2025whitepaper,
  title        = {Bring AI to every corner of your business with ServiceNow and Accenture},
  author       = {{ServiceNow} and {Accenture}},
  howpublished = {White paper},
  year         = {2025},
  month        = nov,
  url          = {https://www.servicenow.com/content/dam/servicenow-assets/public/en-us/doc-type/resource-center/white-paper/servicenow-accenture-ai-whitepaper-final-en-uk-nov-2025.pdf},
  note         = {Accessed 2026-02-11}
}

@misc{opa2024,
  title        = {Open Policy Agent},
  author       = {{Open Policy Agent Contributors}},
  year         = {2024},
  howpublished = {\url{https://openpolicyagent.org/}},
  note         = {Accessed: 2026-02-11}
}

@misc{bedrock2024,
  title        = {Amazon Bedrock Documentation},
  author       = {{Amazon Web Services}},
  year         = {2024},
  howpublished = {\url{https://docs.aws.amazon.com/bedrock/}},
  note         = {Accessed: 2026-02-11}
}

@misc{strands_docs2024,
  title        = {Strands Agents Documentation},
  author       = {{Amazon Web Services}},
  year         = {2024},
  howpublished = {\url{https://strandsagents.com/latest/documentation/docs/}},
  note         = {Accessed: 2026-02-11}
}

@misc{claude_code_overview,
  title        = {Claude Code overview},
  author       = {{Anthropic}},
  howpublished = {\url{https://code.claude.com/docs/en/overview}},
  note         = {Accessed: 2026-02-08}
}

@misc{kiro_cli_intro,
  title        = {Bring Kiro agents to your terminal with Kiro CLI},
  author       = {{Kiro}},
  year         = {2025},
  howpublished = {\url{https://kiro.dev/blog/introducing-kiro-cli/}},
  note         = {Accessed: 2026-02-08}
}

@misc{reinhard2024_genai_it_support,
  title        = {Generative AI in Customer Support Services: A Framework for Augmenting the Routines of Frontline Service Employees},
  author       = {Reinhard, Philipp and others},
  year         = {2024},
  howpublished = {\url{https://papers.ssrn.com/sol3/papers.cfm?abstract_id=4862940}},
  note         = {Accessed: 2026-02-08}
}

@online{MCP,
  author = {Anthropic PBC},
  title = {Introducing the Model Context Protocol},
  year = 2024,
  url = {https://www.anthropic.com/news/model-context-protocol},
  urldate = {2024-11-25}
}

@article{luo2025toward,
  title={Toward edge general intelligence with multiple-large language model (Multi-LLM): architecture, trust, and orchestration},
  author={Luo, Haoxiang and Liu, Yinqiu and Zhang, Ruichen and Wang, Jiacheng and Sun, Gang and Niyato, Dusit and Yu, Hongfang and Xiong, Zehui and Wang, Xianbin and Shen, Xuemin},
  journal={IEEE Transactions on Cognitive Communications and Networking},
  year={2025},
  publisher={IEEE}
}

@article{tallam2025autonomous,
  title={From autonomous agents to integrated systems, a new paradigm: Orchestrated distributed intelligence},
  author={Tallam, Krti},
  journal={arXiv preprint arXiv:2503.13754},
  year={2025}
}

@article{acharya2025agentic,
  title={Agentic ai: Autonomous intelligence for complex goals--a comprehensive survey},
  author={Acharya, Deepak Bhaskar and Kuppan, Karthigeyan and Divya, B},
  journal={IEEE Access},
  year={2025},
  publisher={IEEE}
}

@article{sapkota2025ai,
  title={Ai agents vs. agentic ai: A conceptual taxonomy, applications and challenges},
  author={Sapkota, Ranjan and Roumeliotis, Konstantinos I and Karkee, Manoj},
  journal={arXiv preprint arXiv:2505.10468},
  year={2025}
}

@article{moshkovich2025taming,
  title={Taming uncertainty via automation: Observing, analyzing, and optimizing agentic ai systems},
  author={Moshkovich, Dany and Zeltyn, Sergey},
  journal={arXiv preprint arXiv:2507.11277},
  year={2025}
}

@article{fiorini2025episodic,
  title={Episodic Memory in Agentic Frameworks: Suggesting Next Tasks},
  author={Fiorini, Sandro Rama and Azevedo, Leonardo G and Thiago, Raphael M and de Sousa, Valesca M and Labate, Anton B and da Silva, Viviane Torres},
  journal={arXiv preprint arXiv:2511.17775},
  year={2025}
}

@inproceedings{yao2023react,
  title     = {ReAct: Synergizing Reasoning and Acting in Language Models},
  author    = {Yao, Shunyu and Zhao, Jeffrey and Yu, Dian and Du, Nan and Shafran, Izhak and Narasimhan, Karthik and Cao, Yuan},
  booktitle = {International Conference on Learning Representations},
  year      = {2023},
  url       = {https://arxiv.org/abs/2210.03629}
}

@inproceedings{schick2023toolformer,
  title     = {Toolformer: Language Models Can Teach Themselves to Use Tools},
  author    = {Schick, Timo and Dwivedi-Yu, Jane and Dess{\`i}, Roberto and Raileanu, Roberta and Lomeli, Maria and Hambro, Eric and Zettlemoyer, Luke and Cancedda, Nicola and Scialom, Thomas},
  booktitle = {Advances in Neural Information Processing Systems},
  year      = {2023},
  url       = {https://arxiv.org/abs/2302.04761}
}

@inproceedings{yao2023tree,
  title     = {Tree of Thoughts: Deliberate Problem Solving with Large Language Models},
  author    = {Yao, Shunyu and Yu, Dian and Zhao, Jeffrey and Shafran, Izhak and Griffiths, Thomas L. and Cao, Yuan and Narasimhan, Karthik},
  booktitle = {Advances in Neural Information Processing Systems},
  year      = {2023},
  url       = {https://arxiv.org/abs/2305.10601}
}

@misc{a2a2025spec,
  title        = {{Agent2Agent (A2A) Protocol Specification, Version 1.0}},
  author       = {{Agent2Agent Working Group}},
  year         = {2025},
  howpublished = {\url{https://a2a-protocol.org/latest/specification/}},
  note         = {Open standard for interoperability between AI agents}
}

@article{patil2023gorilla,
  title   = {Gorilla: Large Language Model Connected with Massive APIs},
  author  = {Patil, Shishir G. and Hu, Eric Wallace and others},
  journal = {arXiv preprint arXiv:2305.15334},
  year    = {2023},
  url     = {https://arxiv.org/abs/2305.15334}
}

@article{kirsch2022minimal,
  title   = {Minimizing Epistemic Uncertainty in Deep Learning},
  author  = {Kirsch, Andreas and van Amersfoort, Joost and Gal, Yarin},
  journal = {Advances in Neural Information Processing Systems},
  year    = {2022},
  url     = {https://arxiv.org/abs/2205.06876}
}

@inproceedings{agrawal2024deepuncertainty,
  title     = {Deep Uncertainty Quantification for Decision-Making with Large Language Models},
  author    = {Agrawal, Aditi and Li, Xuechen and Xiong, Caiming and Raghunathan, Aditi},
  booktitle = {International Conference on Learning Representations},
  year      = {2024},
  url       = {https://arxiv.org/abs/2310.14757}
}

@article{gupta2023llmdebugger,
  title   = {LLM-Debugger: Autonomous Debugging with Tool-Integrated Large Language Models},
  author  = {Gupta, Pranay and Suresh, Anirudh and Singh, Damanpreet and others},
  journal = {arXiv preprint arXiv:2311.07480},
  year    = {2023},
  url     = {https://arxiv.org/abs/2311.07480}
}

@inproceedings{santurkar2023whose,
  title     = {Whose Opinions Do Language Models Reflect?},
  author    = {Santurkar, Shashank and Durmus, Esin and Ladhak, Faisal and Liang, Percy and Hashimoto, Tatsunori},
  booktitle = {International Conference on Machine Learning},
  year      = {2023},
  url       = {https://arxiv.org/abs/2303.08944}
}

@inproceedings{madaan2023selfrefine,
  title     = {Self-Refine: Iterative Refinement with Large Language Models},
  author    = {Madaan, Anirudh and Liu, Shuyan and Yazdanbakhsh, Amir and Chen, Xinyun and Lin, Xi Victoria and Zhou, Yuandong},
  booktitle = {Advances in Neural Information Processing Systems},
  year      = {2023},
  url       = {https://arxiv.org/abs/2303.17651}
}

@article{zhu2023repairagent,
  title   = {RepairAgent: Fixing Software Bugs with LLM-Based Autonomous Agents},
  author  = {Zhu, Liang and Ye, Zhe and Li, Yingjun and Sun, Jian},
  journal = {arXiv preprint arXiv:2307.10779},
  year    = {2023},
  url     = {https://arxiv.org/abs/2307.10779}
}

@inproceedings{anokhin2024arigraph,
  title     = {AriGraph: Learning Knowledge Graph World Models with Episodic Memory for LLM Agents},
  author    = {Anokhin, Petr and Semenov, Nikita and Sorokin, Artyom and Evseev, Dmitry and Burtsev, Mikhail and Burnaev, Evgeny},
  booktitle = {Proceedings of IJCAI 2025},
  year      = {2025},
  url       = {https://arxiv.org/abs/2407.04363}
}

@article{cai2025kgllm,
  title   = {Practices, Opportunities and Challenges in the Fusion of Knowledge Graphs and Large Language Models},
  author  = {Cai, Linyue and Yu, Chaojia and Kang, Yongqi and Fu, Yu and Zhang, Heng and Zhao, Yong},
  journal = {Frontiers in Computer Science},
  year    = {2025},
  doi     = {10.3389/fcomp.2025.1590632}
}

@article{laptev2023multivariate,
  title   = {Multivariate Time-Series Anomaly Detection with LLM-Augmented Forecasting},
  author  = {Laptev, Nikolay and others},
  journal = {arXiv preprint arXiv:2310.06745},
  year    = {2023},
  url     = {https://arxiv.org/abs/2310.06745}
}

@article{yu2022graphanomaly,
  title   = {GDN: Graph Neural Networks for Multivariate Time-Series Anomaly Detection},
  author  = {Yu, Jing and others},
  journal = {AAAI},
  year    = {2022},
  url     = {https://arxiv.org/abs/2105.13516}
}

@article{miller2019explanation,
  title   = {Explanation in Artificial Intelligence: Insights from the Social Sciences},
  author  = {Miller, Tim},
  journal = {Artificial Intelligence},
  year    = {2019}
}

@inproceedings{yao2022react,
  title     = {ReAct: Synergizing Reasoning and Acting in Language Models},
  author    = {Yao, Shunyu and Zhao, Jeffrey and Yu, Dian and Du, Nan and Shafran, Izhak and Narasimhan, Karthik and Cao, Yuan},
  booktitle = {International Conference on Learning Representations},
  year      = {2023},
  url       = {https://arxiv.org/abs/2210.03629}
}

@inproceedings{wang2023longmem,
  title     = {Augmenting Language Models with Long-Term Memory},
  author    = {Wang, Weizhe and Yu, Lili and Weston, Jason and Sukhbaatar, Sainbayar},
  booktitle = {Advances in Neural Information Processing Systems},
  year      = {2023},
  url       = {https://arxiv.org/abs/2306.07174}
}

@article{vitui2025empowering,
  title   = {Empowering AIOps: Leveraging Large Language Models for IT Operations Management},
  author  = {Vitui, Arthur and Chen, Tse-Hsun},
  journal = {arXiv preprint arXiv:2501.12461},
  year    = {2025},
  url     = {https://arxiv.org/abs/2501.12461}
}

@article{li2025aiops_survey,
  title   = {AIOps in the Era of Large Language Models},
  author  = {de la Cruz Cabello, M. and others},
  journal = {ACM Computing Surveys},
  year    = {2025},
  doi     = {10.1145/3746635},
  url     = {https://dl.acm.org/doi/10.1145/3746635}
}

@article{zota2025agentic,
  title   = {A Practical Approach to Defining a Framework for Developing an Agentic AIOps System},
  author  = {Zota, Răzvan Daniel and Bărbulescu, Corneliu and Constantinescu, Radu},
  journal = {Electronics},
  volume  = {14},
  number  = {9},
  pages   = {1775},
  year    = {2025},
  doi     = {10.3390/electronics14091775},
  url     = {https://www.mdpi.com/2079-9292/14/9/1775}
}

@article{chen2025aiopslab,
  title   = {AIOPSLAB: A Holistic Framework to Evaluate AI Agents for Enabling Autonomous Clouds},
  author  = {Chen, Yinfang and Shetty, Manish and Somashekar, Gagan and Ma, Minghua and Simmhan, Yogesh and Mace, Jonathan and Bansal, Chetan and Wang, Rujia and Rajmohan, Saravan},
  journal = {arXiv preprint arXiv:2501.06706},
  year    = {2025},
  url     = {https://arxiv.org/abs/2501.06706}
}

@article{gao2023rag_survey,
  title   = {Retrieval-Augmented Generation for Large Language Models: A Survey},
  author  = {Gao, Yunfan and Xiong, Yun and Gao, Xinyu and Jia, Kangxiang and Pan, Jinliu and Bi, Yuxi and Dai, Yi and Sun, Jiawei and Wang, Meng and Wang, Haofen},
  journal = {arXiv preprint arXiv:2312.10997},
  year    = {2023}
}

@article{GEPA,
  title   = {GEPA: Reflective Prompt Evolution Can Outperform Reinforcement Learning},
  author  = {Agrawal, Lakshya A. and Tan, Shangyin and Soylu, Dilara and Ziems, Noah and Khare, Rishi and Opsahl-Ong, Krista and Singhvi, Arnav and Shandilya, Herumb and Ryan, Michael J. and Jiang, Meng and Potts, Christopher and Sen, Koushik and Dimakis, Alexandros G. and Stoica, Ion and Klein, Dan and Zaharia, Matei and Khattab, Omar},
  journal = {arXiv preprint arXiv:2507.19457},
  year    = {2025}
}

@article{ReasoningBank,
  title   = {Scaling Agent Self-Evolving with Reasoning Memory},
  author  = {Ouyang, Sherry and others},
  journal = {arXiv preprint arXiv:2509.25140},
  year    = {2025},
  note    = {Introduces the ReasoningBank memory framework}
}

@inproceedings{Bouzenia2025RepairAgent,
  author    = {Ilias Bouzenia and Marcel B{\"o}hme and Cristian Cadar and others},
  title     = {An Autonomous, LLM-Based Agent for Program Repair},
  booktitle = {ICSE},
  year      = {2025},
  eprint    = {2403.17134}
}

@incollection{brooke1996sus,
  title={SUS-A quick and dirty usability scale},
  author={Brooke, John and others},
  journal={Usability evaluation in industry},
  volume={189},
  number={194},
  pages={4--7},
  year={1996},
  publisher={London, England}
}

@incollection{hart1988nasa,
  title={Development of NASA-TLX (Task Load Index): Results of empirical and theoretical research},
  author={Hart, Sandra G and Staveland, Lowell E},
  booktitle={Advances in psychology},
  volume={52},
  pages={139--183},
  year={1988},
  publisher={Elsevier}
}

@article{jian2000trust,
  title={Foundations for an empirically determined scale of trust in automated systems},
  author={Jian, Jiun-Yin and Bisantz, Ann M and Drury, Colin G},
  journal={International journal of cognitive ergonomics},
  volume={4},
  number={1},
  pages={53--71},
  year={2000},
  publisher={Taylor \& Francis}
}

@article{davis1989tam,
  title={Perceived Usefulness, Perceived Ease of Use and User Acceptance of Information Technology},
  author={Davis, Fred D},
  journal={MIS quarterly},
  year={1989}
}

\clearpage
\section{Appendix}

\subsection{Technology Stack}




\begin{table}[th]
\centering
\small
\begin{tabularx}{\columnwidth}{>{\raggedright\arraybackslash}X X}
\textbf{Layer} & \textbf{Technology} \\
\hline
\\[-6pt]
Agent Framework & AWS Strands (Python) \\
LLM Inference & Claude Opus 4.1\\
Inter-Agent Protocol & Agent2Agent (A2A) \\
Packaging & PyInstaller \\
Observability & OpenTelemetry tracing \\
Policy Enforcement & Open Policy Agent (OPA)\\
Authentication & Enterprise SSO\\
\end{tabularx}
\caption{Core implementation components of VIGIL}
\label{tab:tech_stack}
\end{table}

\subsection{Example Execution Summaries}
After each session, we convert the execution trace into human-readable summaries. A few examples are provided below.

\subsubsection{No Issue Case}

\begin{tcolorbox}[colback=gray!5,
                  colframe=black,
                  boxrule=0.3pt,
                  arc=2pt,
                  breakable]
\footnotesize\ttfamily
\textbf{User Request:}  
"How can I make my computer faster?"\\

\textbf{VIGIL Execution Trace:}\\
1. Invoked system diagnostics:\\
   - CPU utilization: 9\% \\
   - Disk utilization: 0.6\% \\
   - No abnormal processes detected \\
2. Evaluated performance indicators against policy thresholds.  \\
3. Determined system operating within normal parameters.  \\
4. No remediation actions executed. \\

\textbf{Resolution:}  
VIGIL concluded that the system was performing normally and
avoided unnecessary troubleshooting or state-changing actions.
This is considered a successful resolution because it correctly
diagnosed "no issue" rather than applying unnecessary fixes.

\end{tcolorbox}
\newpage
\subsubsection{Failure Case}

\begin{tcolorbox}[colback=gray!5,
                  colframe=black,
                  boxrule=0.3pt,
                  arc=2pt,
                  breakable]
                  \footnotesize\ttfamily
\textbf{User Request:} "my headset mic is not working" \\

\textbf{VIGIL Execution Trace:} \\
1. Opened Sound Control Panel. \\
2. Launched built-in Audio Troubleshooter. \\
3. Attempted to restart audio service (failed – requires admin privileges). \\
4. Detected that the headset was not recognized by the operating system. \\

\textbf{Resolution:} \\
VIGIL determined that the issue could not be resolved remotely. \\
It generated a structured escalation path including: \\
- Manual physical troubleshooting steps. \\
- Request for administrative access to reinstall drivers. \\
- Recommendation to visit IT vending machine for hardware replacement.
\end{tcolorbox}

\subsubsection{Success Case}
\begin{tcolorbox}[colback=gray!5,
                  colframe=black,
                  boxrule=0.3pt,
                  arc=2pt,
                  breakable]
\footnotesize\ttfamily
\textbf{User Request:}  
"My laptop doesn't respond after waking from sleep and requires a hard reboot." \\

\textbf{VIGIL Execution Trace:} \\
Diagnosis Phase: \\
1. Invoked system uptime check to detect abnormal reboot patterns. \\
2. Retrieved relevant enterprise knowledge base articles. \\
3. Verified hardware specifications, driver versions, and pending updates. \\
Remediation Phase: \\
4. Synthesized stepwise remediation plan grounded in retrieved guidance. \\
5. Executed bounded shell commands to adjust power management settings. \\
6. Disabled fast startup configuration. \\
7. Updated advanced power configuration parameters. \\
8. Re-validated system state after each step. \\
Execution Summary: \\
- Total shell commands executed: 11. \\
- Successful command rate: 90.9\%. \\
- All actions executed under policy constraints with user consent. \\

\textbf{Resolution:} \\
VIGIL systematically diagnosed a Windows sleep/wake configuration issue. \\
It applied targeted power-management adjustments under policy control. \\
Continuous verification ensured safe execution and avoided unnecessary or high-risk interventions. \\
\end{tcolorbox}

\onecolumn
\subsection{Evaluation Prompts}

\begin{tcolorbox}[
  breakable,
  colback=gray!5,
  colframe=black,
  boxrule=0.3pt,
  arc=2pt,
  left=6pt,right=6pt,top=6pt,bottom=6pt
]
\footnotesize\ttfamily
==========================\par
EVALUATION PROMPT A: RESPONSE QUALITY\par
==========================\par
\par
You are an expert IT support quality evaluator. Evaluate the following\par
automated IT assistant (VIGIL) response.\par
\par
>> User Issue\par
\{user\_input\}\par
\par
>> VIGIL's Diagnostic Process\par
Tools used: \{tools\_used\}\par
Total cycles: \{total\_cycles\}\par
Duration: \{duration\} seconds\par
\par
Execution trace:\par
\{trace\_summary\}\par
\par
>> VIGIL's Final Response\par
\{response\}\par
\par
---\par
\par
>> Evaluation Criteria\par
\par
Please evaluate the response on the following criteria (1-10 scale,\par
where 10 is excellent):\par
\par
1. Issue Understanding: Did VIGIL correctly understand what the user\par
   was asking about?\par
2. Root Cause Accuracy: Did VIGIL identify the correct root cause or\par
   relevant factors?\par
3. Response Relevance: Is the response directly relevant to the user's\par
   issue?\par
4. Actionability: Are the suggested steps clear, specific, and\par
   actionable?\par
5. Completeness: Does the response address all aspects of the issue?\par
\par
Also assess:\par
6. Resolution Confidence: How confident are you that following VIGIL's\par
   guidance would resolve the issue?\par
   - "high": Very likely to resolve\par
   - "medium": Likely to help but may need follow-up\par
   - "low": Unlikely to fully resolve\par
   - "unable": VIGIL could not provide a resolution\par
\par
Output: JSON with scores and assessment.\par
\end{tcolorbox}

\noindent\small\textbf{Prompt A.} LLM-as-judge prompt used to score response quality for VIGIL interactions.
\label{fig:eval_prompt_quality}

\vspace{1em}

\begin{tcolorbox}[
  breakable,
  colback=gray!5,
  colframe=black,
  boxrule=0.3pt,
  arc=2pt,
  left=6pt,right=6pt,top=6pt,bottom=6pt
]
\footnotesize\ttfamily
==========================\par
EVALUATION PROMPT B: CASE SIMILARITY MATCHING\par
==========================\par
\par
You are an IT support case analyst. Compare the following two IT\par
support cases and determine if they are similar in terms of:\par
1. Issue type (the fundamental problem being addressed)\par
2. Root cause (the underlying cause of the issue)\par
3. Resolution actions (the steps taken to resolve, if available)\par
\par
>> VIGIL Case (Automated IT Assistant)\par
\par
User Issue:\par
\{user\_input\}\par
\par
Final Response:\par
\{response\}\par
\par
>> CGR Case (Human Agent IT Support)\par
\par
Issue Description:\par
\{issue\_description\}\par
\par
Chat Transcript (excerpt):\par
\{chat\_transcript\}\par
\par
Resolution Summary:\par
\{resolution\_Summary\}\par
\par
Root Cause:\par
\{resolution\_RootCause\}\par
\par
Resolution Steps:\par
\{resolution\_ResolutionAgent\}\par
\par
>> Your Task\par
\par
Analyze both cases and provide your assessment in the following JSON\par
format:\par
\{\par
    "is\_similar": true/false,\par
    "confidence": 1-10,\par
    "issue\_type\_match": true/false,\par
    "root\_cause\_match": true/false,\par
    "resolution\_match": true/false,\par
    "reasoning": "Brief explanation of your assessment"\par
\}\par
\end{tcolorbox}

\noindent\small\textbf{Prompt B.} Prompt used to assess similarity between VIGIL-resolved cases and matched human support cases.
\label{fig:eval_prompt_match}

\subsection{Example OPA Policies}
\label{sec:opa_policies}

\begin{tcolorbox}[
  breakable,
  colback=gray!5,
  colframe=black,
  boxrule=0.3pt,
  arc=2pt,
  left=6pt,right=6pt,top=6pt,bottom=6pt
]
\footnotesize\ttfamily
==========================\par
DENY (Blocked - Never Execute)\par
==========================\par
\par
1. Command: format C:\par
   Description: Attempts to format the C: drive, erasing the system disk\par
   including the OS, user files, and installed programs. Catastrophically\par
   destructive and irreversible.\par
\par
2. Command:\par
   deny[msg] \{\par
       contains(input.command, "net user")\par
       contains(input.command, "/add")\par
       msg := "Blocked: User creation command detected"\par
   \}\par
   Description: Blocks creation of new user accounts that could establish\par
   unauthorized persistent access.\par
\par
3. Command:\par
   deny[msg] \{\par
       lower\_cmd := lower(input.command)\par
       contains(lower\_cmd, "netsh")\par
       contains(lower\_cmd, "firewall")\par
       contains(lower\_cmd, "delete")\par
       msg := "Blocked: Firewall rule deletion command detected"\par
   \}\par
   Description: Prevents deletion of firewall rules that protect the system\par
   from network threats. Captures syntactic variations to avoid bypass.\par
\par
==========================\par
ALLOW (Execute Immediately)\par
==========================\par
\par
1. Command: Get-Process\par
   Description: Lists running processes with CPU/memory usage.\par
   Read-only diagnostic command.\par
\par
2. Command: systeminfo\par
   Description: Displays system configuration details including hardware,\par
   OS version, and network settings. Read-only diagnostic command.\par
\par
==========================\par
WARN (Requires User Approval)\par
==========================\par
\par
1. Command: Stop-Process -Name chrome\par
   Description: Terminates Chrome and may close unsaved work.\par
   Requires user confirmation.\par
\par
2. Command: winget install 7zip\par
   Description: Installs software via Windows Package Manager.\par
   System-modifying but legitimate and reversible; requires approval.\par
\end{tcolorbox}

\noindent\small Illustrative OPA-based command classification examples used in the deployed system.

\twocolumn
\subsection{Expanded SUS Item-Level Analysis}

\begin{figure}[t]
\centering
\includegraphics[width=0.9\columnwidth]{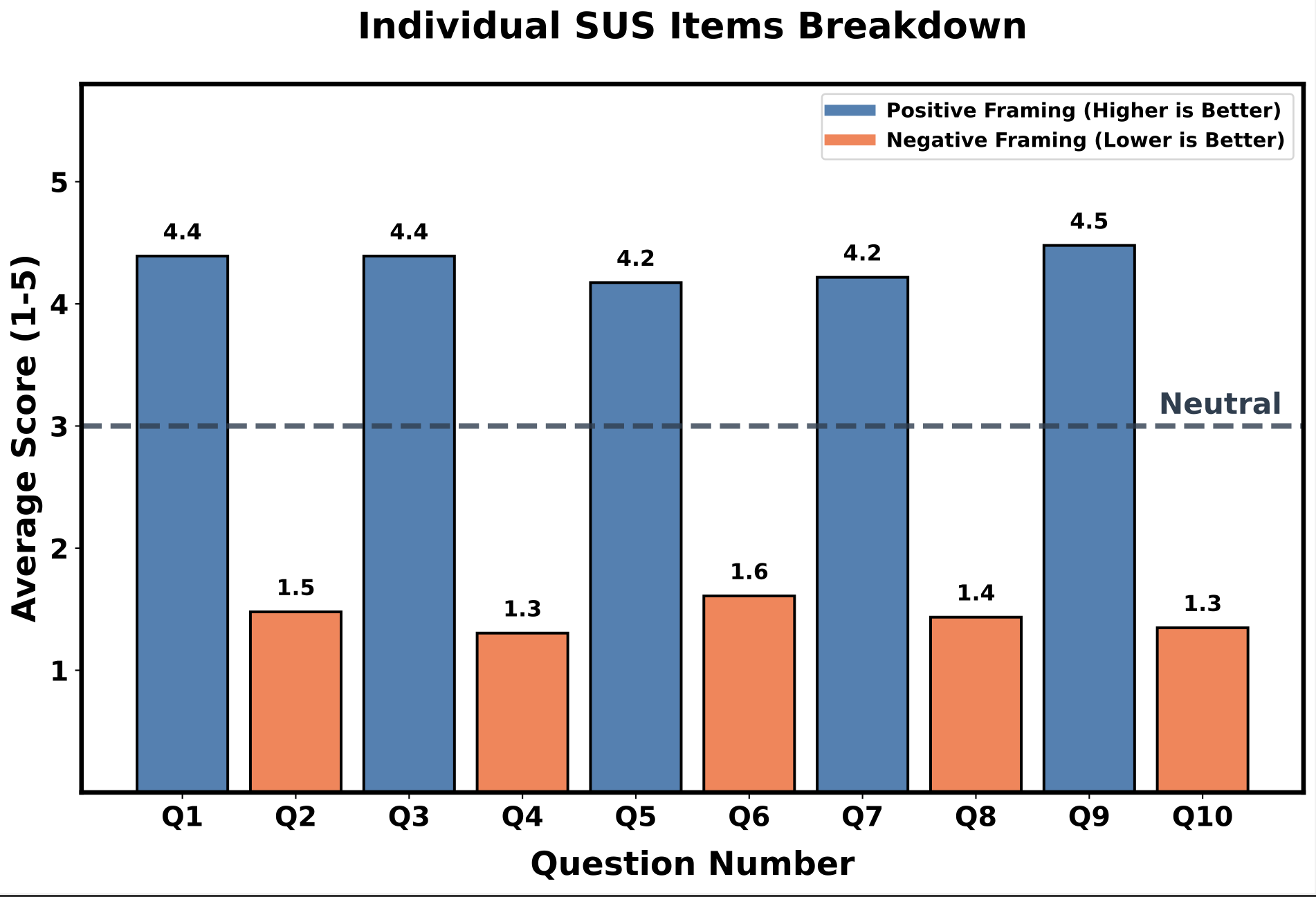}
\caption{Item-level SUS score breakdown (1--5 scale). Positive-framed items (higher is better) and negative-framed items (lower is better) are shown separately. The dashed line indicates the neutral midpoint (3.0).}
\label{fig:sus_item_breakdown}
\end{figure}

To provide greater transparency into the System Usability Scale (SUS) results reported in the main text, we include an item-level breakdown of all ten SUS questions in Figure \ref{fig:sus_item_breakdown}. Positive-framed items (Q1, Q3, Q5, Q7, Q9) are shown directly, where higher scores indicate stronger agreement and better usability perception. Negative-framed items (Q2, Q4, Q6, Q8, Q10) are shown in their original form (lower is better), allowing inspection of perceived complexity and friction. The neutral midpoint (3.0) is included for reference. This breakdown highlights consistency across usability dimensions and helps interpret the aggregate SUS score of 86.2. The questions are enumerated below for reference:

\begin{enumerate}[label={}, leftmargin=0.5em]
    \item Q1. I think that I would like to use this system frequently.
    \item Q2. I found the system unnecessarily complex.
    \item Q3. I thought the system was easy to use.
    \item Q4. I think that I would need the support of a technical person to be able to use this system.
    \item Q5. I found the various functions in this system were well integrated.
    \item Q6. I thought there was too much inconsistency in this system.
    \item Q7. I would imagine that most people would learn to use this system very quickly.
    \item Q8. I found the system very cumbersome to use.
    \item Q9. I felt very confident using the system.
    \item Q10. I needed to learn a lot of things before I could get going with this system.
\end{enumerate}

\end{document}